\documentclass{article} %
\usepackage{iclr2027_conference,times}
\iclrfinalcopy

\usepackage{amsmath,amsfonts,bm}

\def\eqref#1{equation~\ref{#1}}

\def\1{\bm{1}}

\DeclareMathAlphabet{\mathsfit}{\encodingdefault}{\sfdefault}{m}{sl}
\SetMathAlphabet{\mathsfit}{bold}{\encodingdefault}{\sfdefault}{bx}{n}

\DeclareMathOperator*{\argmax}{arg\,max}
\DeclareMathOperator*{\argmin}{arg\,min}

\usepackage{hyperref}
\usepackage{url}
\usepackage{microtype}
\usepackage{graphicx}
\usepackage{subfigure}
\usepackage{booktabs}
\usepackage{float}

\usepackage{amsmath}
\usepackage{amssymb}
\usepackage{mathtools}
\usepackage{amsthm}

\title{Physically Interpretable Basin Selection in Discrete-Variable Origami Design}

\author{Lucy Liu$^{1}$, Leon M. Kamp$^{1}$, Katia Bertoldi$^{1}$, \& L. Mahadevan$^{1,2,3}$ \\
$^{1}$John A. Paulson School of
Engineering and Applied Sciences, \\ $^{2}$Department of Physics, $^{3}$Department of Organismic and
Evolutionary Biology \\
Harvard University, Cambridge, MA 02138, USA \\
\texttt{\{lliu,lmahadev\}@g.harvard.edu, \{lkamp,bertoldi\}@seas.harvard.edu}
}

\begin{document}

\maketitle

\begin{abstract}

Automatic differentiation has enabled rapid progress on continuous-variable mechanical design problems, but gradient-based optimization still struggles with discrete variables. The standard approach is to optimize using a continuous relaxation of the choices, then commit to the nearest discrete design at the end. Why this procedure selects unfavorable basins is poorly understood. Here, we introduce a practical, interpretable design problem inspired by origami, in which automatic differentiation is used to design shapes with a target functional behavior. Each origami sheet has a fixed geometry, and a bilevel optimization framework tunes which one of four discrete hinge types to place at each fold. Baseline gradient-based design settles into basins with a large discretization gap: relaxed optimization solves 97\% of runs, but the final, discrete solutions succeed only half the time. We find that the gap can be relieved with a surprisingly simple modification. An initialization prior uniformly biased toward the stiff, flat hinge option boosts the final solve fraction from 49\% to 69\%. Applying the prior adds no additional compute and leaves the optimization process unchanged. Compared to the baseline, sheets produced with the prior have greater geometric compatibility and mechanical stiffness. Directly penalizing frustration and floppiness in the outer loss function produces an even greater improvement than the prior does, raising the solve fraction to 78\% and indicating that these properties help distinguish the unfavorable basins. Our results demonstrate that loss landscapes for mechanical design can have physically meaningful basins that are not robust to discretization and cause solves to fail. Targeting these basins' physical signatures, even with just an initialization bias, helps avoid them.

\quad

\end{abstract}

\section{Introduction}
Origami and other art forms have inspired new materials, known as \textit{metamaterials}, which can exhibit programmed abilities like morphing into desired shapes or directing energy to target points. Metamaterial design is often posed as solving for a network of variables that specifies how to fabricate the material. Each variable has a physical meaning, such as the \textit{x}-coordinate of a vertex or the length of a specific edge, and variables representing nearby elements are generally coupled. A major challenge is finding solutions that exhibit the target behavior while respecting geometric constraints, especially when the design variables are discrete.

One promising approach is using gradient-based optimization to minimize an objective function with respect to the design variables. The objective function can combine loss terms associated with arbitrarily many goals and constraints, as long as they are expressed in a differentiable manner. It can even encode simulations of physical processes. Furthermore, the gradient can now be computed quickly, without taking any derivatives by hand, using automatic differentiation software \citep{paszke2019pytorch}. Gradient-based design via automatic differentiation has been used to optimize continuous design variables for self-assembling particles \citep{king2024programming},  inflation-based walking robots \citep{matthews2023efficient}, and origami's sibling, kirigami \citep{bordiga2024automated}. However, fabrication options are often discrete, not continuous, and discrete variables are a known challenge for gradient-based methods \citep{chen2025generative}. The standard workaround, optimizing with a continuous relaxation \citep{bendsoe1989optimal} and projecting back to the discrete options at the end, can converge to suboptimal solutions \citep{gillman2019discovering, stolpe2001trajectories}. Why this happens is poorly understood.

Here, we introduce a functional origami problem: designing ``flowguiding" sheets that use their topography to direct fluid flow to target sink locations, like watersheds in a mountain range. Flowguide design has potential applications in robotics \citep{wehner2016integrated} and in autonomous experimentation platforms \citep{damir2026autonomous}. Our task is to produce the desired functionality by selecting one of four discrete hinge types to place at each origami crease. We formulate the problem as a bilevel optimization with a continuous relaxation. Because origami structures can be visualized and even fabricated, our design problem is an interpretable testbed for dissecting the challenges gradient-based design faces in constrained optimization and discrete type selection. %

Following \citet{kamp2026bistable}, our origami sheets have a fixed topology and geometry, meaning the lengths, locations, and connectivity of creases do not change. However, each crease is assigned one of four hinge types, which defines its rest angle and bending energetics. The optimization variable is the map $a$ of hinge type assignments.

After introducing related work in Section~\ref{section:related_work}, we define the flowguide design problem and optimization model in Sections~\ref{section:background} and~\ref{section:model}. In Section~\ref{section:baseline}, we characterize the performance of the baseline solve, which suffers from a discretization gap. The relaxed solve nearly always solves the flowguiding problem, but the final, discrete hinge assignments succeed only 49\% of the time.

Next, in Section~\ref{section:initialization_prior}, we show that an initialization prior biased toward the stiff, flat hinge type raises the solve fraction to 69\%. This is surprising for two reasons. First, conventional optimization wisdom favors unbiased initializations with maximum optionality. Second, existing work on improving discrete design, which is discussed in Section~\ref{section:related_work}, has typically adjusted the optimization process, not the initialization.

We find that origami sheets designed using the prior have higher stiffness and geometric compatibility, and their 3D shapes change less during the discretization step. To further probe the relationship between these physical signatures and the increased solve fraction, in Section~\ref{section:penalties} we try replacing the prior with direct penalties on floppiness and bending frustration in the outer loss function. This alternative intervention raises the solve fraction even further, to 78\%. These results suggest that optimization from the prior descends into different and physically interpretable loss basins. Identifying the characteristics distinguishing these basins and steering optimization toward them improves the chance that a solve succeeds. 

Altogether, our work improves the success rate of discrete inverse origami design and sheds light on the loss landscape of a constrained mechanical design problem. The relaxed loss landscape has many local minima. However, only some correspond to solutions that can be successfully discretized and fabricated. To encourage finding these solutions, one typically alters the loss landscape (by adding terms to the objective function) or the optimization method. We have now found that an initialization prior, which leaves the landscape and optimizer untouched and corresponds to starting the search at a specific part of the landscape, can also significantly improve the success of gradient-based design. This indicates that the loss landscape has a robust region where nearby minima are more likely to be solutions that survive discretization.

\section{Related Work}\label{section:related_work}

Decades of research have explored how to fold flat sheets into desired 2D and 3D shapes, including animals, vases, and even vibrant patterns formed using different colors on the sheet's front and back \citep{demaine1999folding, demaine2007geometric, dudte2016programming, dudte2021additive}. Building on these advances in understanding static shape, interest has now expanded to how origami interacts with the world around it \citep{nain2024tunable}. 

Incorporating interactions and competing forces into inverse design often involves a bilevel problem formulation. An outer level updates design parameters. An inner level solves for the system's equilibrium state under the current design, accounting for all the interacting elements. Such nested design has been demonstrated with continuous design
variables in metamaterials and computational fabrication \citep{ren2022umbrella}. 

Of prior work, the most similar to ours is \citet{gillman2019discovering}, which uses a nested optimization to identify which candidate edges should be creased (as opposed to left flat) to produce origami actuators. 
\citet{gillman2019discovering} and \citet{shende2021bayesian} ultimately find that gradient-based optimization produces suboptimal designs compared to genetic algorithms and Bayesian optimization. Other works on discrete materials design have found success in settings where the inner problem is convex \citep{stegmann2005discrete, dold2023differentiable}, or turned to using reinforcement learning \citep{geiger2023automatingrigid} and random forests \citep{chen2025generative}. Our study focuses on improving the success rate of gradient-based optimization, which encodes physics directly in its objective functions and does not require any training data or combinatorial search. Our inner solve is also highly multistable, a known property of origami \citep{waitukaitis2015origami, grasinger2026lagrangian}.

Bilevel optimization is also used in machine learning for hyperparameter tuning \citep{bengio2000gradient, maclaurin2015gradient} and for automatically designing network architectures. DARTS \citep{liu2018darts}, an optimization framework similar to ours, uses an approximate bilevel solve to select one operation from a discrete set of candidates to assign to each connection in an architecture. DARTS is known to have a number of failure modes; \citet{zela2020understanding} find that DARTS struggles when loss curvature is very high, and that interventions for avoiding such solutions improve generalization. \citet{ruochenwang2021dartspt} develop new selection criteria for discretizing the relaxed solution to the final design. In the origami setting, strong mechanical coupling between optimization variables produces new failure modes not applicable to DARTS.

\section{Background}\label{section:background}
\begin{figure}
    \centering
    \includegraphics[width=1\linewidth]{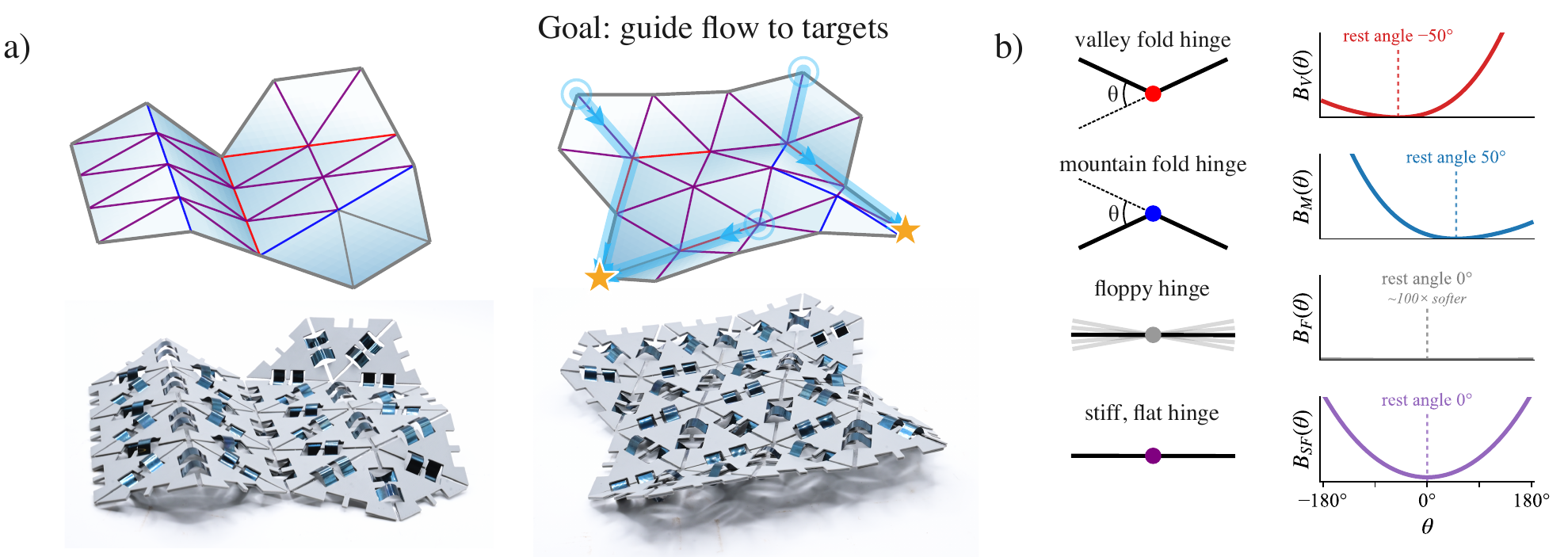}
    \caption{Origami sheets are designed by selecting which of four hinge types to place at each interior edge. \textbf{a)} Left: different hinge types exhibit different behaviors, such as mountain or valley folds. Right: The objective is to select hinge types so that fluid on the sheet will flow to target points (stars). Several flow paths are shown in light blue, and deeper shading indicates lower elevation. Each example is accompanied by a photo of a matching experimental origami sheet (details in Appendix~\ref{si_experimental_images}). \textbf{b)} The four hinge options and their bending energy functions. The angle $\theta$ is measured so that the dashed line is $0^\circ$ and counterclockwise is positive.}  %
    \label{fig:F1_setting}
\end{figure}

\subsection{Origami Sheet}
Our study focuses on designing origami for directing fluid flow to target locations. In particular, we model the experimental origami sheets developed in \cite{kamp2026bistable}, which fabricates origami sheets using rigid panels connected by programmable hinges. The panels correspond to uncreased facets of the origami, and the hinges to potential folds. We use the panel geometry shown in Fig.~\ref{fig:F1_setting}a throughout this work. The sheet has $n_v = 22$ vertices and $n_h = 35$ interior hinges to be assigned types. The panels are congruent $50-65-65$ degree isosceles triangles.

\subsection{Hinge Types}
 Flexible spring steel shims placed between adjacent panels are used to program the behavior of each hinge (see \cite{kamp2026bistable} for more details). Let $\mathcal{T} = \{\mathrm{V}, \mathrm{M}, \mathrm{F}, \mathrm{SF}\}$ be the set of permitted hinge types. There are four types of hinges, for which sketches and bending energy profiles are shown in Fig.~\ref{fig:F1_setting}b:
\begin{itemize}
    \item Type $\mathrm{V}$: A valley fold hinge (red) has a rest angle of $\theta_0 = -50.2^\circ$. See Fig.~\ref{fig:F1_setting}b for a depiction of how $\theta$ is measured. 
    \item Type $\mathrm{M}$: A mountain fold hinge (blue) has a rest angle of $\theta_0 = 50.2^\circ$. It is physically identical to the valley fold hinge, but the shims are buckled in the opposite direction.
    \item Type $\mathrm{F}$: A floppy hinge (gray) has no extra shims. Its rest angle is $0^\circ$, meaning it prefers to lie flat, but its bending energy is about 100$\times$ lower than the other shims'.
    \item Type $\mathrm{SF}$: A stiff, flat hinge (purple) has two shims buckled in opposite directions. Their opposing symmetrical forces cause the hinge to prefer a flat state ($\theta_0 = 0^\circ$). Its bending energy is of a magnitude comparable to that of the valley and mountain fold hinges. 
\end{itemize}

Torque-angle functions were obtained by fitting degree-5 polynomials to data from finite element simulations of the hinge structures. Integrating the torque-angle functions then produced bending energy functions $B_V(\theta), B_M(\theta), B_F(\theta),$ and $B_{SF}(\theta)$ for each hinge type, which are plotted in Fig.\ref{fig:F1_setting}c and each take in the angle $\theta$ and return the hinge's bending energy. The approximate behavior is that each hinge has a preferred rest angle, and its bending energy increases as its angle deviates from the rest angle. See the supplemental information of \cite{kamp2026bistable} for more detailed information, along with the exact functions. \cite{kamp2026bistable} studies experimental sheets that can be reprogrammed to form different 3D shapes by pressing the shims into different buckling configurations to change whether they are mountain, valley, or stiff-flat.

\section{Model}\label{section:model}
\begin{figure}
    \centering
    \includegraphics[width=1\linewidth]{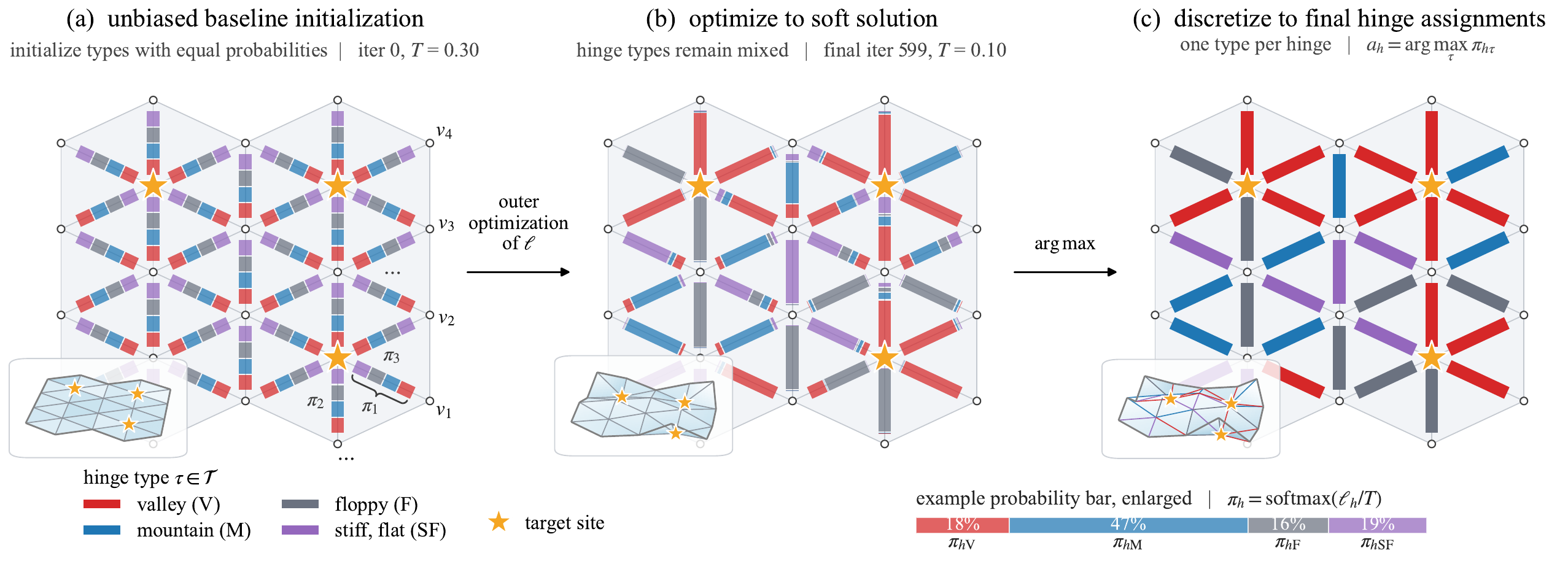}
    \caption{The baseline continuous relaxation, illustrated. Insets display the current 3D equilibrium shape. \textbf{a)} Each hinge is initialized to equal logits on all four types, plus symmetry-breaking noise. Raw logit vectors $\ell_h$ are converted to hinge probabilities $\pi_h$ using $\pi_h = \text{softmax}\left(\frac{\ell}{T}\right)$. \textbf{b)} During optimization, gradient steps update the hinge logits $\ell_h$. Annealing and a regularization term sharpen the mixtures so that one hinge type leads. \textbf{c)} After the final optimization iteration, the soft, mixed solution is discretized so that one final type is assigned per hinge.}
    \label{fig:si_continuous_relaxation_demo}
\end{figure}

\subsection{Problem Statement}
The shape of an origami sheet is determined by an assignment $a \in \{\mathrm{V}, \mathrm{M}, \mathrm{F}, \mathrm{SF}\}^{n_h}$ of one hinge type to each hinge. The task is to find $a$ so that the final 3D shape, a local minimum of the sheet energy $E$, guides simulated fluid flow to specific target points on the sheet. For an example, see Fig.~\ref{fig:F1_setting}a, where flow is guided to targets marked with gold stars.

\subsection{Continuous relaxation}

To learn the discrete hinge assignments using gradient steps, we employ a standard continuous relaxation. Gradient design works by making small updates to the variable representing each hinge. When the only hinge options are valley, mountain, floppy, or stiff-flat, small updates are not possible. The relaxation treats each hinge as a mixture of the different types, which allows the gradient design to take small steps making a hinge ``a little more valley'' or ``a little less mountain''. When the small-step updates are complete, the mixed solution is projected back to the closest discrete design.

During the relaxed optimization, hinges can be a mixture of different types, which is visualized in Fig.~\ref{fig:si_continuous_relaxation_demo}a and b. Each hinge $h$ has a logits vector $\ell_h$ containing a score for each hinge type. A softmax parametrized by temperature $T$ maps the logits to a distribution $\pi_h = \text{softmax}(\ell_h/T)$, whose components $\pi_{h\tau}$ express how strongly hinge type $\tau$ contributes to hinge $h$'s energy. The matrix $\Pi$ refers to all the $\pi_h$ vectors stacked together.

For the baseline solve, logits are initialized to the same value for each type, plus some symmetry-breaking noise, so the optimization begins with approximately equal probabilities across all hinge types (Fig.~\ref{fig:si_continuous_relaxation_demo}a). The specific initialization value for each logit is $\ell_{h\tau} = 0.02u_{h\tau}$, where $u_{h\tau}\sim \text{Unif}(0,1)$ is drawn independently for each hinge and each type.

$T$ is annealed from 0.3 to 0.1 over the first 37 of 600 outer iterations, sharpening the mixtures toward one-hot distributions (Fig.~\ref{fig:si_continuous_relaxation_demo}b). After the gradient updates are complete, argmax is used to project the soft solution to a discrete assignment $a$ with one type per hinge: $a_h = \argmax_{\tau \in \mathcal{T}} \pi_{h\tau}$ (Fig.~\ref{fig:si_continuous_relaxation_demo}c). A final equilibrium solve and flow simulation are run using $a$ and used to determine the reported result. Because the discretized sheets tend to be more frustrated, the final equilibrium solve ends with an extra 500 inner steps at a $10\times$ smaller learning rate to help the result converge.

\begin{figure}
    \centering
    \includegraphics[width=0.8\linewidth]{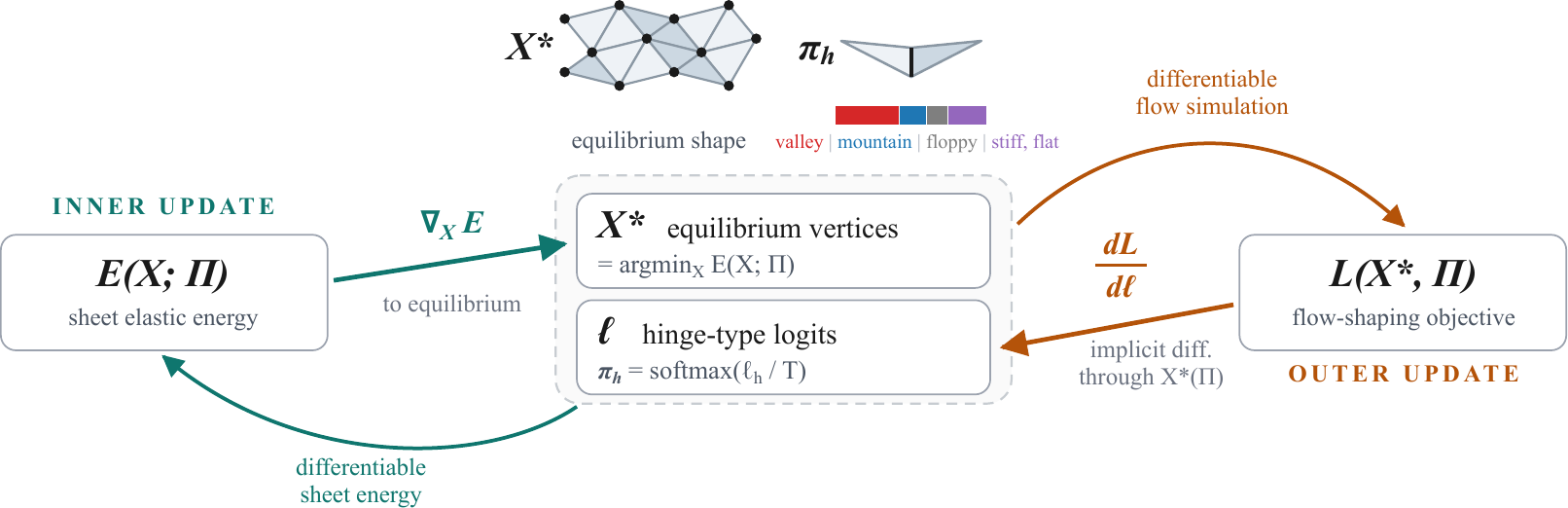}
    \caption{The bilevel optimization structure used to find the Fig.\ref{fig:si_continuous_relaxation_demo}b soft solution. During the inner solve, the 3D shape $X$ of the sheet is updated to a local minimum $X^*$ of the sheet elastic energy. During the outer solve, the hinge type logits $\ell$ are updated to minimize the objective loss, in turn reshaping the sheet's energy landscape.
    }
    \label{fig:F1_optimization}
\end{figure}

\subsection{Inner loop: sheet physics}

During the inner loop, hinge type scores are held constant while the vertex positions are updated to find the shape of the relaxed sheet. 

\textbf{Sheet energy.} The elastic energy function $E$ models the origami sheet as a series of axial and rotational springs used to represent stretching and bending energy, as has been previously used to study both origami \citep{waitukaitis2015origami} and other thin-sheet systems \citep{liang2009shape}. An elastic sheet's rest shape is a stable equilibrium of its total energy $E$. Then if $X \in \mathbb{R}^{n_v \times 3}$ collects all vertex positions, the rest shape is a vertex configuration $X^*$ that locally minimizes $E(X; a)$ for discrete assignments or $E(X; \Pi)$ for relaxed assignments; details are provided in Appendix~\ref{si_sheet_energy}. To predict the final 3D shape resulting from hinge assignments $a$, an inner solve relaxes $X$ from a fixed near-flat starting configuration $X_0$ to a local minimum of $E$.

 \textbf{Optimization loop.} We compute the sheet energy in PyTorch, and during the update step, automatic differentiation is used to take $\frac{\partial E}{\partial X}$. For the updates, we use the Adam optimizer with learning rate $3\times 10^{-4}$ and up to 50000 steps. The inner solve concludes early, before its maximum cap, if the energy plateaus or if the force residual (the mean energy gradient magnitude per vertex) falls below $10^{-3}$. 
 
 This energy formulation and a similar optimization process were shown in \cite{kamp2026bistable} to produce equilibrated sheet shapes that match experiments with physical sheets. Because our work relaxes thousands of different, often frustrated, sheets and assumes each one has reached equilibrium, our hyperparameters and solve details are not exactly the same as in \cite{kamp2026bistable}. However, we check that for each simulation-experiment match in \cite{kamp2026bistable}, the inner solve used in this work reproduces the match.

\subsection{Outer loop: design for functional behavior}

In the outer update, sheet shape is held fixed at the equilibrated $X^*$, and the hinge assignments are updated to minimize the outer loss $\mathcal{L}$, with gradients taken through the equilibrium shape $X^*(\Pi)$ via the implicit function theorem. In each step of the outer loop, a full inner solve of up to 50000 inner steps is run to obtain the relaxed sheet shape $X^*$ corresponding to the current hinge assignments. The outer loop runs for 600 hinge update steps and also uses the Adam optimizer.

\textbf{Differentiable flow simulation. }  

After the inner loop finds equilibrium shape $X^*(a)$ for a set of hinge assignments, the outer loop runs a differentiable simulation of flow on the sheet to predict where the flow would collect. The premise of the simulation is that flow moves downhill, and that because the origami sheet contains straight edges and planar faces, this property can be encapsulated in a Markov transition matrix $P_F$ once $X^*$ is known. Then updating the flow distribution is akin to multiplying by the matrix (Appendix~\ref{si_flow_simulation}).
For the experiments in this paper, the flow starting distribution is uniform across all sites. The transition matrix is applied for 10 updates. The resulting matrix-vector product represents the distribution of where the flow collects on the sheet. The outer loss is the KL divergence between the target site distribution and the resulting distribution on the final sheet, $\mathcal{L}_{\text{flow}}(X^*(\Pi))$, plus an entropy penalty that penalizes indecisiveness. 

\begin{equation}\label{eq:baseline_outer_loss}
  \mathcal{L}(\Pi) =
  \underbrace{\mathrm{KL}\!\left(p^{\text{target}} \,\middle\|\, p(X^*(\Pi))\right)}_{\mathcal{L}_{\text{flow}}}
  \;+\;
  \frac{\lambda_{\mathcal{H}}}{n_h}\sum_{h} \mathcal{H}(\pi_h),
  \qquad
  \mathcal{H}(\pi_h) = -\sum_{\tau \in \mathcal{T}} \pi_{h\tau}\log \pi_{h\tau},
\end{equation}
where $n_h$ is the number of hinges and $\lambda_{\mathcal{H}} = 0.297$. In our implementation, the KL term is divided by the number of sites $|S| = 99$ (PyTorch batchmean reduction), so the entropy weight appears in our code as $0.297/99 = 0.003$.

\textbf{Hypergradient.} 
To find the hypergradient describing how to update the hinge logits to minimize $\mathcal{L}$, we use the implicit function theorem and the adjoint method. Similar numerical approaches have been employed in both materials science \citep{bordiga2024automated} and in hyperparameter optimization for neural networks \citep{bengio2000gradient}. See Appendix~\ref{si_hypergradient} for details.

\textbf{Experiments.}
The optimization is evaluated on a set of 24 different flowguiding problems on the origami sheet (Fig.~\ref{si_flow_problems}). The number of targets $k$ varies from one to three. For a problem with $k$ goals, the target flow distribution is to deliver $\frac{1}{k}$ of the total flow to each goal. Each problem is run using 12 different random seeds for the initialization noise. 

We deem a run successfully solved if a nontrivial amount of flow ($\geq 10\%$ of the desired flow at each target) is delivered to all target sites on the final discrete sheet. A method's \textit{solve fraction} or \textit{discrete solve fraction} is the fraction of its runs that succeed across all problems and seeds. In some sections, we also discuss the \textit{soft solve fraction}, which refers to this same test run on the soft, relaxed design at the end of the gradient updates, before each hinge is discretized to a single type.

\section{Baseline Optimization}\label{section:baseline}
\begin{figure}
    \centering
    \includegraphics[width=0.8\linewidth]{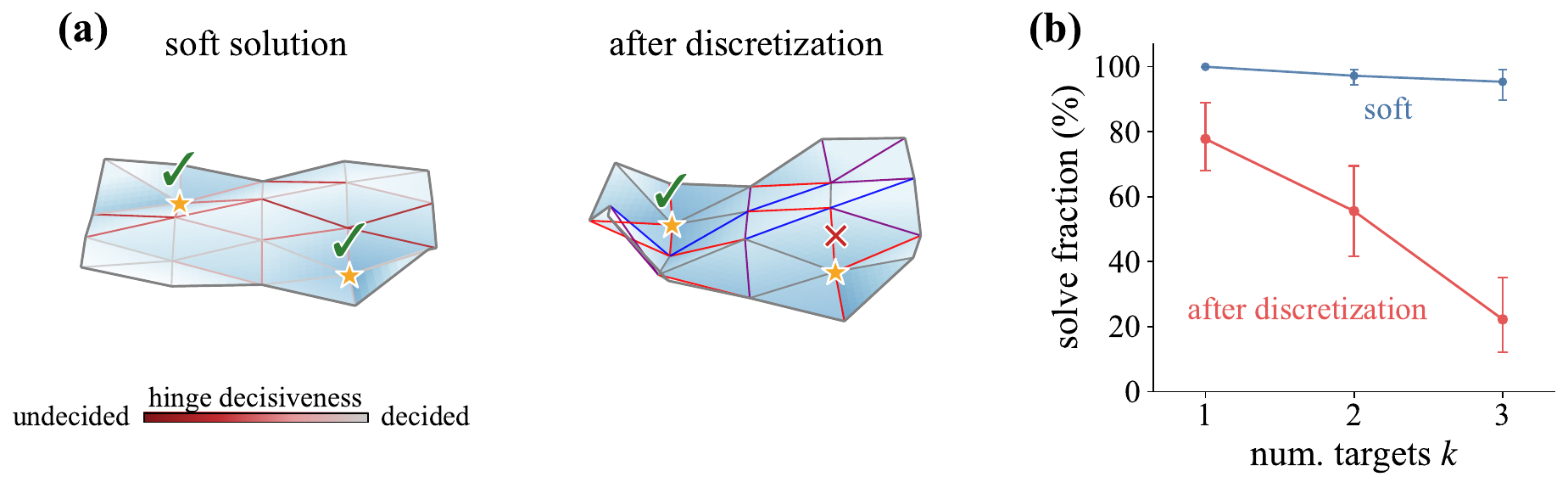}
    \caption{Baseline optimization performance. \textbf{a)} Example where the baseline solve fails at discretization. Target points are marked with stars. A checkmark indicates successful flow direction to a target. \textbf{b)} The soft solution solves nearly all problems, and failures arise after discretization. Solve fraction drops as the number of targets $k$ increases. Error bars show 95\% bootstrap confidence intervals obtained by resampling problems within each $k$. } 
    \label{fig:F2_baseline}
\end{figure}

The naive baseline relaxation has a 49\% solve fraction. The solve fraction drops as the number of targets $k$ increases, from 78\% for $k=1$ problems to 56\% for $k=2$ and 22\% for $k=3$. Further analysis reveals that during the soft optimization phase with relaxed hinges, baseline optimization almost always solves the problem, as seen in Fig.~\ref{fig:F2_baseline}b. The failures arise when the soft, mixed hinges are discretized to assign each hinge one final type. For some soft solutions, discretizing the hinges changes the sheet's equilibrium shape enough that the solve no longer succeeds. For an example of a sheet failing at discretization, see Fig.~\ref{fig:F2_baseline}a.

For hinges entirely committed to one hinge type, the bending energy is not affected by the discretization step. The more a hinge splits its probability score across multiple hinge types at discretization time, the more the argmax discretization alters its behavior. Let a hinge be ``strongly decided" if a single hinge type is assigned at least $90\%$ of its probability. At the end of the baseline method's outer updates, only $65\%$ of hinges are strongly decided.

Finally, we assess the local compatibility of the discrete hinge assignments found by the baseline solve. We call the configuration of six hinges adjacent to each interior vertex a ``vertex star". Due to geometric constraints, not all vertex star hinge assignments are compatible. By relaxing an isolated vertex star to its energy-minimizing shape and observing how much frustration remains, we can determine if it is incompatible (see Appendix~\ref{si_incompatibility} for a more precise definition). In the baseline solve's final discrete designs, 12\% of vertex stars are incompatible, and 65\% of designs have at least one incompatible vertex star.

\section{A Stiff, Flat Initialization Prior}\label{section:initialization_prior}

\begin{figure}
    \centering
    \includegraphics[width=0.7\linewidth]{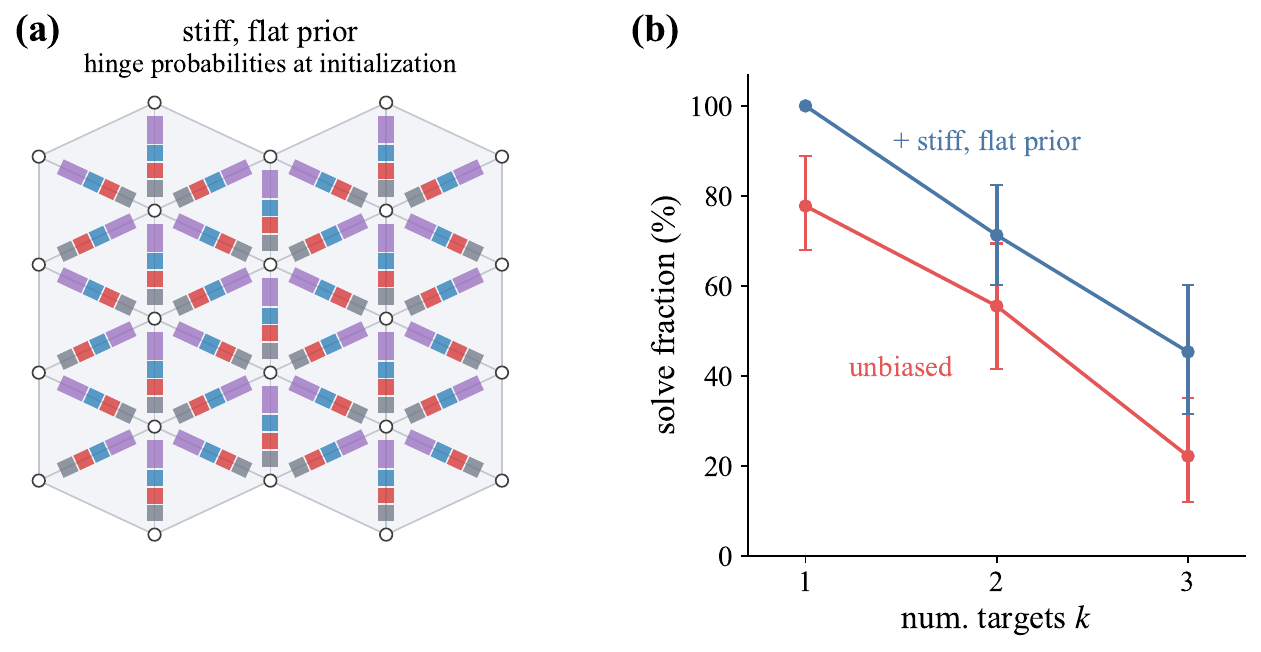}
    \caption{Initialization prior. \textbf{a)} Hinge initialization probabilities for the stiff, flat prior. Purple represents the probability assigned to the SF hinge type. \textbf{b)} Baseline vs. prior solve fraction for different values of $k$. Error bars show 95\% bootstrap confidence intervals obtained by resampling problems within each $k$. }
    \label{fig:F3_interventions}
\end{figure}

An origami sheet comprised entirely of stiff, flat hinges remains flat and has no frustration. Motivated by how frequently incompatible configurations appear in the baseline solve, we implement an initialization bias toward stiff, flat hinges. While the baseline solve initializes hinge probabilities equally across types (plus symmetry-breaking noise), the biased initialization adds an additional 0.15 to each $\mathrm{SF}$ logit. Each hinge's $\pi_{h\mathrm{SF}}$ then begins at about $35\%$ instead of $25\%$, which is illustrated in Fig.~\ref{fig:F3_interventions}a.

Adding the stiff, flat prior to the baseline optimization raises the solve fraction from 49\% to 69\%. The prior also reduces incompatibility, producing origami designs where 4\% of vertex stars are incompatible and 26\% of designs contain at least one incompatible star.

\begin{figure}
    \centering
    \includegraphics[width=0.9\linewidth]{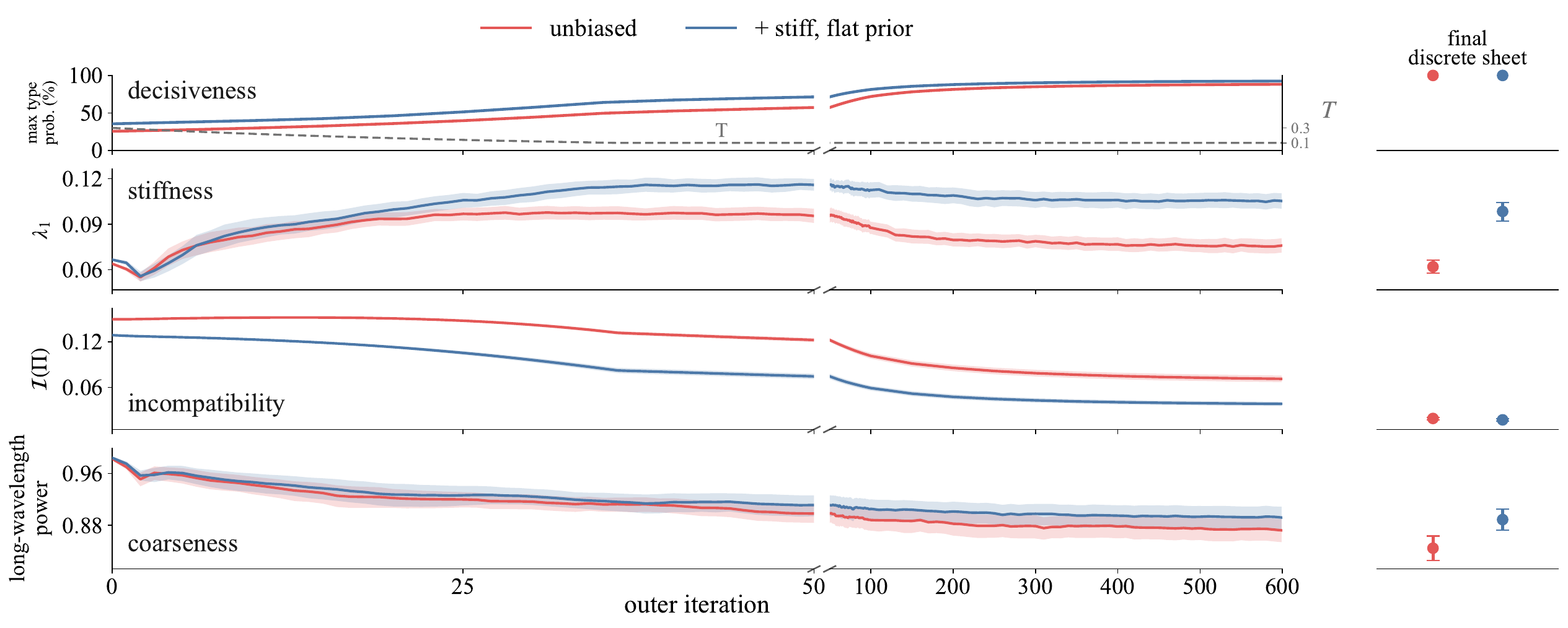}
    \caption{Left: Soft optimization trajectories of baseline optimization vs. stiff-flat prior. Right: the same metrics after discretization. We measure decisiveness using the average per-hinge $\max(\pi_{h\tau})$, stiffness using the smallest eigenvalue of the Hessian $H(E)(X^*; \Pi)$, geometric incompatibility using the excess energy per vertex star when relaxed in isolation $\mathcal{I}(\Pi)$ (Appendix~\ref{si_incompatibility}), and coarseness using the fraction of the spectral power (obtained via Fourier transform of the sheet's $z$-coordinates) in long-wavelength frequencies (Appendix~\ref{si_fourier}). Bands and error bars show 95\% bootstrap confidence intervals obtained by resampling problems within each $k$.}
    \label{fig:F4_trajectories}
\end{figure}

To further understand how the stiff, flat prior changes the optimization, in Fig.~\ref{fig:F4_trajectories} we track an array of metrics over each sheet's optimization trajectory.  
\begin{itemize}
    \item Decisiveness: Highest probability assigned to a hinge type, averaged over all hinges.
    \item Stiffness: $\lambda_1$, the lowest Hessian eigenvalue excluding the six rigid-body modes, which is similarly used by \citet{ren2022umbrella}.
    \item Incompatibility: The average excess elastic energy $\mathcal{I}(\Pi)$ in a vertex star due to incompatibility between the hinges; see Appendix~\ref{si_incompatibility} for details.
    \item Coarseness: Percentage of the spectral power present in long-wavelength modes of the Fourier transform (Appendix~\ref{si_fourier}). Coarseness indicates a smoother final shape.
\end{itemize}

The prior finds sheets which are stiffer, geometrically coarser, and contain more compatible hinge combinations. The effect on compatibility is visible from iteration 0, while the effects on stiffness and coarseness emerge and widen later during optimization. Hinges are also more decisive early on in the solve, although the gap in decisiveness narrows by the end of optimization. Adding the prior, which represents changing the optimization's initialization point in hinge-assignment search space, thus moves the solve to a basin with distinct and physically interpretable signatures. 

The stiffness and compatibility encouraged by the prior are qualities that intuitively help a sheet remain robust to the discretization step. Incompatible stars have hinges held away from their desired rest angles, so they are frustrated and store excess bending energy. Even small changes to hinge energies due to discretization could release this stored energy and cause the sheet to relax to a different position, changing its shape and flow paths. Meanwhile, stiffness is a desirable quality because it means the sheet resists changes to its current shape. Indeed, the prior finds designs whose shapes change less during discretization (Table~\ref{tab:penalty}).

\section{Loss Penalties}\label{section:penalties}

Next, we probe whether the increased stiffness and compatibility have a causal or merely correlational relationship with the improved solve fraction. Instead of applying an initialization prior, we experiment with adding penalties for floppiness and incompatibility to the baseline solve's outer loss (Equation~\ref{eq:baseline_outer_loss}). Specifically, we add regularizer terms penalizing the excess bending frustration of the equilibrium sheet and the prevalence of  floppy ($\mathrm{F}$-type) hinges in the current design (details in Appendix~\ref{si_extra_penalties}). 

\begin{table}[t]
\centering
\caption{Adding loss penalties. The shape change measures the mean $|\Delta\theta|$ per hinge between the soft and discrete equilibrium shapes, in degrees, and the other metrics match their earlier definitions from Sections~\ref{section:model} and~\ref{section:baseline}. The bottom block tests removing the floppy type entirely from the set of hinge type options. Each row corresponds to 288 runs (24 flowguide problems with 12 seeds each). The $p$-value is from the clustered McNemar test of the discrete solve fraction against the stiff, flat prior result in the top block. Runs are matched by problem and seed, and the problems serve as the different clusters. Using this test, the unbiased baseline solve fraction is significantly different from every other row's, $p < 0.01$. }

\label{tab:penalty}
\small
\setlength{\tabcolsep}{4pt}
\begin{tabular}{lrrrrr}
\toprule
 & \textbf{soft solution} & \multicolumn{2}{c}{\textbf{after discretizing}} & \textbf{\% stars} & \textbf{shape}\\
\cmidrule(lr){3-4}
 & \textbf{solve frac. (\%)} & \textbf{solve frac. (\%)} & \textbf{$p$ vs prior} & \textbf{incompat.} & \textbf{change ($^\circ$)}\\
\midrule
\multicolumn{6}{l}{\textit{no penalty}}\\
\hspace{0.5em}unbiased baseline & 97.2 & 48.6 & 0.001 & 12.4 & 12.0\\
\hspace{0.5em}stiff, flat prior & 95.8 & 68.8 & -- & 4.3 & 4.6\\
\midrule
\multicolumn{6}{l}{\textit{penalties}}\\
\hspace{0.5em}floppiness & 97.6 & 72.6 & 0.34 & 20.3 & 6.6\\
\hspace{0.5em}frustration & 93.8 & 30.6 & $<$0.001 & 0.2 & 11.7\\
\hspace{0.5em}frustration + floppiness & 97.6 & 78.1 & 0.04 & 2.8 & 2.9\\
\midrule
\multicolumn{6}{l}{\textit{F removed from $\mathcal{T}$}}\\
\hspace{0.5em}unbiased baseline & 97.2 & 63.2 & 0.23 & 39.5 & 9.5\\
\hspace{0.5em}stiff, flat prior & 99.0 & 74.0 & 0.20 & 7.8 & 4.5\\
\hspace{0.5em}frustration penalty & 99.3 & 84.4 & 0.004 & 4.5 & 3.1\\
\bottomrule
\end{tabular}
\end{table}

Results are shown in Table~\ref{tab:penalty}. Adding loss penalties for floppiness or floppiness + frustration increases the baseline solve fraction just like the stiff-flat prior does. In fact, penalizing both floppiness and frustration produces an increase in solve fraction and compatibility over the prior. As expected, penalizing floppiness increases the stiffness of final solutions. Penalizing frustration increases compatibility, but selects more floppy hinges and exhibits a reduced solve fraction (lower than the baseline solve) unless an additional intervention is used to reduce floppiness. 

Increased solve fraction compared to the baseline is also observed when we try removing the floppy hinge type from the set of options. The most effective intervention removes the floppy type option and adds a frustration penalty, raising the solve fraction to 84\%. These results demonstrate that targeting stiffness and compatibility suffice to reproduce and even surpass the solve fraction improvement seen with the prior.

\section{Conclusion}

In this work, we sought to understand why gradient-based design struggles to optimize discrete variables. Using the origami flowguide setting as a practical and interpretable testbed, we found evidence that only some basins in the loss landscape are robust to discretization. The robust basins have physically interpretable characteristics, namely stiffness and geometric compatibility. Targeting these properties with either an initialization bias or a loss penalty helps gradient-based design find solutions in the robust basins. Compared to a baseline solve fraction of 49\%, the initialization prior and penalty interventions respectively boost the flowguide design success rate to 69\% and 78\%. They are also both simple for practitioners to implement. A natural next step is to explore the loss landscapes of additional discrete design settings. When does a design problem have favorable regions in its loss landscape, and how can the most effective priors for accessing them be found?

\newpage

\bibliography{iclr2027_conference}

\begin{thebibliography}{29}
\providecommand{\natexlab}[1]{#1}
\providecommand{\url}[1]{\texttt{#1}}
\expandafter\ifx\csname urlstyle\endcsname\relax
  \providecommand{\doi}[1]{doi: #1}\else
  \providecommand{\doi}{doi: \begingroup \urlstyle{rm}\Url}\fi

\bibitem[Bends{\o}e(1989)]{bendsoe1989optimal}
Martin~P Bends{\o}e.
\newblock Optimal shape design as a material distribution problem.
\newblock \emph{Structural Optimization}, 1\penalty0 (4):\penalty0 193--202, 1989.

\bibitem[Bengio(2000)]{bengio2000gradient}
Yoshua Bengio.
\newblock Gradient-based optimization of hyperparameters.
\newblock \emph{Neural Computation}, 12\penalty0 (8):\penalty0 1889--1900, 2000.

\bibitem[Bordiga et~al.(2024)Bordiga, Medina, Jafarzadeh, B{\"o}sch, Adams, Tournat, and Bertoldi]{bordiga2024automated}
G.~Bordiga, E.~Medina, S.~Jafarzadeh, C.~B{\"o}sch, R.~P. Adams, V.~Tournat, and K.~Bertoldi.
\newblock Automated discovery of reprogrammable nonlinear dynamic metamaterials.
\newblock \emph{Nature Materials}, 23\penalty0 (11):\penalty0 1486--1494, September 2024.
\newblock ISSN 1476-4660.
\newblock \doi{10.1038/s41563-024-02008-6}.

\bibitem[Chen et~al.(2025)Chen, Sun, Lee, Portela, and Chen]{chen2025generative}
Wei Chen, Rachel Sun, Doksoo Lee, Carlos~M Portela, and Wei Chen.
\newblock Generative inverse design of metamaterials with functional responses by interpretable learning.
\newblock \emph{Advanced Intelligent Systems}, 7\penalty0 (6):\penalty0 2400611, 2025.

\bibitem[Damir et~al.(2026)Damir, Delgado-Licona, Demello, and Abolhasani]{damir2026autonomous}
Suyash Damir, Fernando Delgado-Licona, Andrew Demello, and Milad Abolhasani.
\newblock Autonomous microfluidic labs: progress and prospects.
\newblock \emph{Lab on a Chip}, 26\penalty0 (5):\penalty0 1014--1024, 2026.

\bibitem[Demaine \& O'Rourke(2007)Demaine and O'Rourke]{demaine2007geometric}
Erik~D Demaine and Joseph O'Rourke.
\newblock \emph{Geometric folding algorithms: linkages, origami, polyhedra}.
\newblock Cambridge university press, 2007.

\bibitem[Demaine et~al.(1999)Demaine, Demaine, and Mitchell]{demaine1999folding}
Erik~D Demaine, Martin~L Demaine, and Joseph~SB Mitchell.
\newblock Folding flat silhouettes and wrapping polyhedral packages: New results in computational origami.
\newblock In \emph{Proceedings of the fifteenth annual symposium on Computational geometry}, pp.\  105--114, 1999.

\bibitem[Dold \& van Egmond(2023)Dold and van Egmond]{dold2023differentiable}
Dominik Dold and Derek~Aranguren van Egmond.
\newblock Differentiable graph-structured models for inverse design of lattice materials.
\newblock \emph{Cell Reports Physical Science}, 4\penalty0 (10), 2023.

\bibitem[Dudte et~al.(2016)Dudte, Vouga, Tachi, and Mahadevan]{dudte2016programming}
Levi~H Dudte, Etienne Vouga, Tomohiro Tachi, and Lakshminarayanan Mahadevan.
\newblock Programming curvature using origami tessellations.
\newblock \emph{Nature Materials}, 15\penalty0 (5):\penalty0 583--588, 2016.

\bibitem[Dudte et~al.(2021)Dudte, Choi, and Mahadevan]{dudte2021additive}
Levi~H Dudte, Gary~PT Choi, and L~Mahadevan.
\newblock An additive algorithm for origami design.
\newblock \emph{Proceedings of the National Academy of Sciences}, 118\penalty0 (21):\penalty0 e2019241118, 2021.

\bibitem[Geiger et~al.(2023)Geiger, Martinkus, Richter, and Wattenhofer]{geiger2023automatingrigid}
Jeremia Geiger, Karolis Martinkus, Oliver Richter, and Roger Wattenhofer.
\newblock Automating rigid origami design.
\newblock In \emph{Proceedings of the Thirty-Second International Joint Conference on Artificial Intelligence}, IJCAI '23, 2023.
\newblock ISBN 978-1-956792-03-4.
\newblock \doi{10.24963/ijcai.2023/645}.
\newblock URL \url{https://doi.org/10.24963/ijcai.2023/645}.

\bibitem[Gillman et~al.(2019)Gillman, Fuchi, and Buskohl]{gillman2019discovering}
Andrew~S Gillman, Kazuko Fuchi, and Philip~R Buskohl.
\newblock Discovering sequenced origami folding through nonlinear mechanics and topology optimization.
\newblock \emph{Journal of Mechanical Design}, 141\penalty0 (4):\penalty0 041401, 2019.

\bibitem[Grasinger et~al.(2026)Grasinger, Gillman, and Buskohl]{grasinger2026lagrangian}
Matthew Grasinger, Andrew Gillman, and Philip Buskohl.
\newblock Lagrangian approach to origami vertex analysis: Multistability.
\newblock \emph{arXiv preprint arXiv:2609.13724}, 2026.

\bibitem[Kamp et~al.(2026)Kamp, Liu, McRitchie, Rouchouse, Mazor, Farhadi, Mahadevan, and Bertoldi]{kamp2026bistable}
Leon~M. Kamp, Lucy Liu, Ella McRitchie, Damien Rouchouse, Orel Mazor, Davood Farhadi, L.~Mahadevan, and Katia Bertoldi.
\newblock Reprogrammable origami through bistable buckled hinges.
\newblock \emph{arXiv preprint arXiv:2609.28790}, 2026.
\newblock URL \url{https://arxiv.org/abs/2609.28790}.

\bibitem[King et~al.(2024)King, Du, Zhu, Schoenholz, and Brenner]{king2024programming}
Ella~M King, Chrisy~Xiyu Du, Qian-Ze Zhu, Samuel~S Schoenholz, and Michael~P Brenner.
\newblock Programming patchy particles for materials assembly design.
\newblock \emph{Proceedings of the National Academy of Sciences}, 121\penalty0 (27):\penalty0 e2311891121, 2024.

\bibitem[Liang \& Mahadevan(2009)Liang and Mahadevan]{liang2009shape}
Haiyi Liang and Lakshminarayanan Mahadevan.
\newblock The shape of a long leaf.
\newblock \emph{Proceedings of the National Academy of Sciences}, 106\penalty0 (52):\penalty0 22049--22054, 2009.

\bibitem[Liu et~al.(2019)Liu, Simonyan, and Yang]{liu2018darts}
Hanxiao Liu, Karen Simonyan, and Yiming Yang.
\newblock {DARTS}: Differentiable architecture search.
\newblock In \emph{International Conference on Learning Representations}, 2019.

\bibitem[Maclaurin et~al.(2015)Maclaurin, Duvenaud, and Adams]{maclaurin2015gradient}
Dougal Maclaurin, David Duvenaud, and Ryan Adams.
\newblock Gradient-based hyperparameter optimization through reversible learning.
\newblock In \emph{International conference on machine learning}, pp.\  2113--2122. PMLR, 2015.

\bibitem[Matthews et~al.(2023)Matthews, Spielberg, Rus, Kriegman, and Bongard]{matthews2023efficient}
David Matthews, Andrew Spielberg, Daniela Rus, Sam Kriegman, and Josh Bongard.
\newblock Efficient automatic design of robots.
\newblock \emph{Proceedings of the National Academy of Sciences}, 120\penalty0 (41):\penalty0 e2305180120, 2023.

\bibitem[Nain et~al.(2024)Nain, Marzin, and Ramananarivo]{nain2024tunable}
Rishabh Nain, Tom Marzin, and Sophie Ramananarivo.
\newblock Tunable drag drop via flow-induced snap-through in origami.
\newblock \emph{Journal of Fluid Mechanics}, 1000:\penalty0 A8, 2024.

\bibitem[Paszke et~al.(2019)Paszke, Gross, Massa, Lerer, Bradbury, Chanan, Killeen, Lin, Gimelshein, Antiga, et~al.]{paszke2019pytorch}
Adam Paszke, Sam Gross, Francisco Massa, Adam Lerer, James Bradbury, Gregory Chanan, Trevor Killeen, Zeming Lin, Natalia Gimelshein, Luca Antiga, et~al.
\newblock Pytorch: An imperative style, high-performance deep learning library.
\newblock \emph{Advances in Neural Information Processing Systems}, 32, 2019.

\bibitem[Ren et~al.(2022)Ren, Kusupati, Panetta, Isvoranu, Pellis, Chen, and Pauly]{ren2022umbrella}
Yingying Ren, Uday Kusupati, Julian Panetta, Florin Isvoranu, Davide Pellis, Tian Chen, and Mark Pauly.
\newblock Umbrella meshes: elastic mechanisms for freeform shape deployment.
\newblock \emph{ACM Transactions on Graphics}, 41\penalty0 (4), July 2022.
\newblock ISSN 0730-0301.
\newblock \doi{10.1145/3528223.3530089}.
\newblock URL \url{https://doi.org/10.1145/3528223.3530089}.

\bibitem[Shende et~al.(2021)Shende, Gillman, Yoo, Buskohl, and Vemaganti]{shende2021bayesian}
Sourabh Shende, Andrew Gillman, David Yoo, Philip Buskohl, and Kumar Vemaganti.
\newblock Bayesian topology optimization for efficient design of origami folding structures.
\newblock \emph{Structural and Multidisciplinary Optimization}, 63\penalty0 (4):\penalty0 1907--1926, 2021.

\bibitem[Stegmann \& Lund(2005)Stegmann and Lund]{stegmann2005discrete}
Jan Stegmann and Erik Lund.
\newblock Discrete material optimization of general composite shell structures.
\newblock \emph{International Journal for Numerical Methods in Engineering}, 62\penalty0 (14):\penalty0 2009--2027, 2005.

\bibitem[Stolpe \& Svanberg(2001)Stolpe and Svanberg]{stolpe2001trajectories}
Mathias Stolpe and Krister Svanberg.
\newblock On the trajectories of penalization methods for topology optimization.
\newblock \emph{Structural and Multidisciplinary Optimization}, 21\penalty0 (2):\penalty0 128--139, 2001.

\bibitem[Waitukaitis et~al.(2015)Waitukaitis, Menaut, Chen, and Van~Hecke]{waitukaitis2015origami}
Scott Waitukaitis, R{\'e}mi Menaut, Bryan Gin-ge Chen, and Martin Van~Hecke.
\newblock Origami multistability: from single vertices to metasheets.
\newblock \emph{Physical Review Letters}, 114\penalty0 (5):\penalty0 055503, 2015.

\bibitem[Wang et~al.(2021)Wang, Cheng, Chen, Tang, and Hsieh]{ruochenwang2021dartspt}
Ruochen Wang, Minhao Cheng, Xiangning Chen, Xiaocheng Tang, and Cho-Jui Hsieh.
\newblock Rethinking architecture selection in differentiable {NAS}.
\newblock In \emph{International Conference on Learning Representations (ICLR)}, 2021.
\newblock URL \url{https://arxiv.org/abs/2108.04392}.

\bibitem[Wehner et~al.(2016)Wehner, Truby, Fitzgerald, Mosadegh, Whitesides, Lewis, and Wood]{wehner2016integrated}
Michael Wehner, Ryan~L Truby, Daniel~J Fitzgerald, Bobak Mosadegh, George~M Whitesides, Jennifer~A Lewis, and Robert~J Wood.
\newblock An integrated design and fabrication strategy for entirely soft, autonomous robots.
\newblock \emph{Nature}, 536\penalty0 (7617):\penalty0 451--455, 2016.

\bibitem[Zela et~al.(2020)Zela, Elsken, Saikia, Marrakchi, Brox, and Hutter]{zela2020understanding}
Arber Zela, Thomas Elsken, Tonmoy Saikia, Yassine Marrakchi, Thomas Brox, and Frank Hutter.
\newblock Understanding and robustifying differentiable architecture search.
\newblock In \emph{International Conference on Learning Representations}, 2020.
\newblock URL \url{https://openreview.net/forum?id=H1gDNyrKDS}.

\end{thebibliography}
\bibliographystyle{iclr2027_conference}

\appendix
\section{Appendix}

\subsection{Experimental Images}\label{si_experimental_images}
The physical sheet shown in photos is a reprogrammable sheet of the type introduced by \citet{kamp2026bistable}, whose hinges can be switched between different types. At the hinges, spring-steel shims of length 20 mm are compressed to fit gaps of length 15 mm, causing the shims to buckle into arches. Note that the sample's hinges can only be reprogrammed between types M, V, and SF, so for the physical samples, each of the floppy hinges in the examples (3 total across both examples) has been substituted with the best matching non-floppy hinge type. 

\subsection{Sheet Energy}\label{si_sheet_energy}

$X_0$ is the flat $z=0$ base origami sheet with a slight curved $z$-perturbation added to it, so $z_i = cx_i^2$, where $c=5 \times 10^{-3}$ scales the perturbation magnitude, and the starting sheet is symmetric in $x$.

First, we introduce the energy function for discrete assignments $a$. A bending energy term captures the frustration in the hinges from being held at angles other than their rest angle. Let $k_b$ be the bending weight, $\theta$ store the hinge angles, and $a_h$ describe the type assigned to hinge $h$. Then $E_{\text{bend}}(X) = k_b \sum_{\text{hinge}~h} B_{a_h}(\theta_h)$.

A stretching energy term captures the constraint that the triangular origami panels are stiff and do not change shape. To represent the property that panels are rigid and do not stretch, stretching energy is penalized using a $10^7$ higher weight than bending energy. Edges are modeled as linear springs $e_1, e_2, ..., n_e$ with current length $l^1, l^2, ..., l^{n_e}$ and rest length $l^1_0, l^2_0, ..., l^{n_e}_0$. Then the stretching energy term is $E_{\text{stretch}}(X) = k_s \sum_{\text{edge}~e} (l^e - l^e_0)^2 $. 

Our total sheet energy is thus
\begin{equation}
    E(X; a) = k_s E_{\text{stretch}}(X) + k_b E_{\text{bend}}(X; a)
\end{equation}

We use $k_s = 10^4, k_b = 10^{-3}$.

The sheet is entirely free and $E$ is invariant to rotations and translations. No boundary conditions are used to fix the sheet's global position. 

During the relaxed optimization steps, we use a relaxed sheet energy $E(X; \Pi)$ which, for each hinge $h$, simply weights the bending energy corresponding to each hinge type according to $\pi_{h\tau}$. The hinge assignments only affect the bending energy, so $E_{\text{stretch}}$ remains untouched:
$$E(X; \Pi) = k_s E_{\text{stretch}}(X) + k_b \sum_{h}\sum_{\tau \in \mathcal{T}} \pi_{h\tau}B_{\tau}(\theta_h)$$

\subsection{Differentiable Flow Simulation}\label{si_flow_simulation}
In the flow simulation, the sheet is represented by an unweighted, undirected flow adjacency graph with sites at vertices, face centroids, and edge midpoints. Flow is allowed to travel between neighbors in the flow graph. Each vertex site is neighbors with the sites representing the vertex's adjacent vertices, faces, and edges.

Using $X^*$ and the flow adjacency graph, we generate a transition matrix $P_F$ where $P_F[i,j]$ stores the probability of transitioning from site $i$ to site $j$ and is nonzero only for neighboring $i,j$. Thus a row of the matrix describes the probability distribution for where fluid currently at site $i$ will go next. 

For each site $i$, a score is generated expressing the flow's preference to move to each neighbor. The primary scoring criteria is that higher scores are assigned to moves with steeper downhill slope. Uphill moves are given a small nonzero score to maintain some gradient signal. Finally, each row is normalized into a probability distribution: for any $i$, $\sum_j P_F[i,j] = 1$.

The differentiable flow simulation captures some key characteristics of physical flow, but it has some behavioral differences as well who do not match reality. For example, the simulated flow distribution will ``hug" the border of the sheet and continue traveling downhill to neighbors instead of falling off the sheet. We also note that the 10 update steps we apply are usually not enough to fully converge to the stationary distribution. This ensures that configurations are rewarded not only for directing flow to the correct locations, but also for directing flow quickly by producing well-defined folds (as opposed to shallow folds).

The flow guiding problem is sensitive, and small differences in tilt angle can completely change where the flow travels to.

\subsection{Hypergradient Computation}\label{si_hypergradient}

Recall that $X^*(\ell)$ is a local minimum of $E(X; \ell)$. (For this section, assume $X$, $X^*$, and $\ell$ have been flattened into vectors.) During the outer optimization, we seek a soft solution
$$\ell^* = \argmin_{\ell}\; \mathcal{L}\bigl(X^*(\ell);\ell\bigr).$$

We want to update the hinge logits $\ell$ in the direction that will decrease the outer loss $\mathcal{L}$ using $\frac{d\mathcal{L}}{d\ell}$. To find $\frac{d\mathcal{L}}{d\ell}$, we first apply multivariate chain rule:
\begin{equation}\label{eq:hypergrad_chain}
    \frac{d\mathcal{L}}{d\ell}
    = \frac{\partial \mathcal{L}}{\partial X}\cdot\frac{dX^*}{d\ell}
    + \frac{\partial \mathcal{L}}{\partial \ell},
\end{equation}

where all partial derivatives are evaluated at $X = X^*(\ell)$. The terms $\frac{\partial \mathcal{L}}{\partial X}$ and $\frac{\partial \mathcal{L}}{\partial \ell}$ can be obtained directly with automatic differentiation calls. To compute $\frac{d X^*}{d \ell}$ without differentiating through tens of thousands of inner solve steps, observe that $X^*$ locally minimizing $E(X; \ell)$ implies $\frac{\partial E}{\partial X} = 0$ at $X^*$. We further assume that the Hessian is invertible at $X^*$ once rigid-body modes are excluded. Under this assumption, the implicit function theorem gives that $X^*(\ell)$ is differentiable and that $\frac{\partial E}{\partial X}\bigl(X^*(\ell);\ell\bigr) = 0$ holds not just for the current $\ell$, but at all $\ell$ in a neighborhood around it. This allows us to differentiate both sides of the identity with respect to $\ell$, yielding

\begin{align*}
    \frac{\partial E}{\partial X}\bigl(X^*(\ell);\ell\bigr) &= 0 
    \\ \implies \frac{d}{d\ell}\left[\frac{\partial E}{\partial X}(X^*(\ell);\ell)\right] = \frac{d}{d \ell} 0 &= 0 &&\text{differentiate both sides}
    \\ \implies \frac{\partial}{\partial X}\frac{\partial E}{\partial X}\cdot \frac{d X^*}{d \ell} + \frac{\partial}{\partial \ell} \frac{\partial E}{\partial X} &= 0 &&\text{expand LHS via chain rule}
    \\ \implies \frac{d X^*}{d \ell} &= -\left(\frac{\partial^2 E}{\partial X^{2}}\right)^{-1} \frac{\partial^2 E}{\partial \ell \partial X} &&\text{rearrange}
\end{align*}

Note that $\frac{\partial^2 E}{\partial X^{2}}$ is the Hessian $H$ of $E$ with respect to $X$. After rewriting $\frac{d X^*}{d \ell}$, we are now able to compute $\frac{d\mathcal{L}}{d\ell}$ using more automatic differentiation calls and linear algebra operations.

In practice, our hypergradient uses a modified version of the exact Hessian. First, the components of the Hessian which are attributable to rigid-body transformations (rotations and translations) are projected out, which is needed to make the Hessian invertible. When computing the Hessian for updating the hinge types, we also assume that the origami panels, which were meant to be stiff, have remained unstrained. That is, their edge lengths and their rest lengths are equal, so the stretching energy and stretching forces at $X^*$ are zero. These modifications remove contributions to $H$ that are not meant to be controlled by the hinge types (global pose and stretching strain in edges) and that can distort the hypergradient when left in.

\subsection{Set of Flow Problems}\label{si_flow_problems}
\begin{figure}[H]
    \centering
    \includegraphics[width=0.85\linewidth]{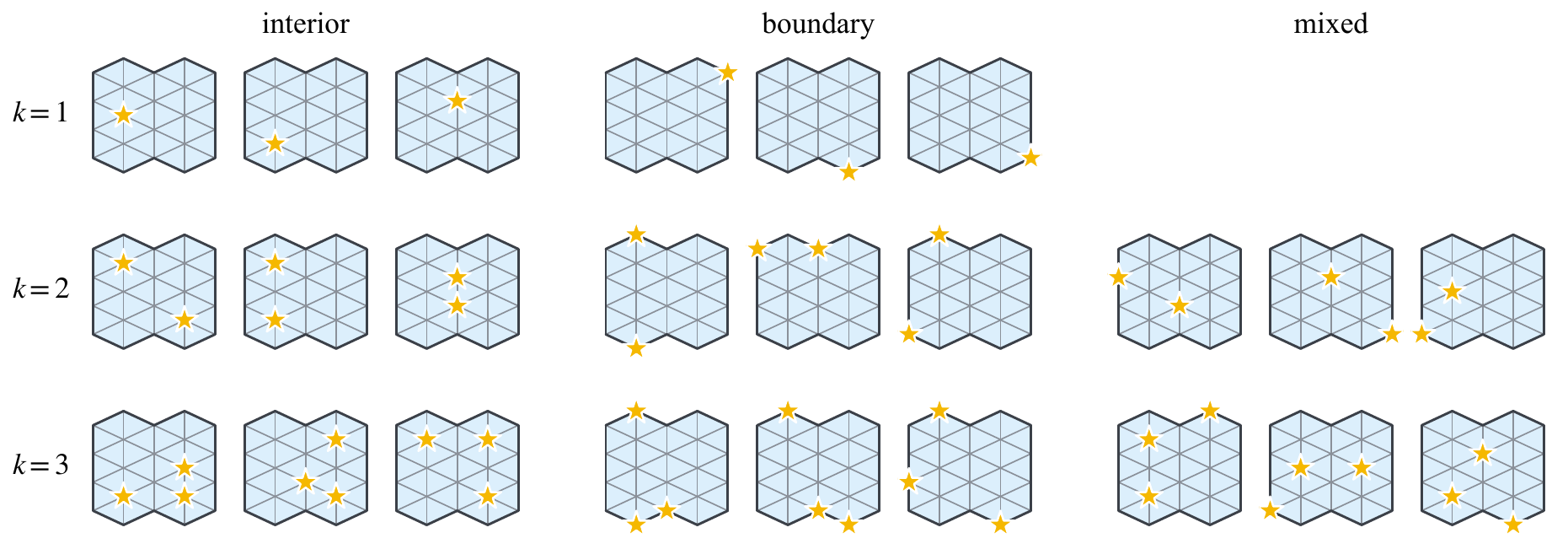}
    \caption{Optimization approaches are tested on 24 different problems.}
    \label{fig:si_problem_set}
\end{figure}

We run each optimization method on a set of 24 flow guiding problems, shown in Fig.~\ref{fig:si_problem_set}, with 6 random seeds per problem. 

\subsection{Measuring Vertex Star Incompatibility}\label{si_incompatibility}

 Some vertex stars are perfectly compatible, meaning a 3D configuration exists where all hinges can take their rest angle. For instance, a star with all-flat hinges is perfectly satisfied in the flat state, where all hinges lie flat as desired. A star with four flat hinges and two valley fold hinges placed across from each other is perfectly satisfied when it creases in half along the valley hinges. However, most vertex stars are not perfectly compatible. 
 
 By running an elastic energy minimization (inner solve) on an isolated vertex star, we can find its energy-minimizing shape and measure its internal frustration using the excess energy remaining there. Specifically, let star $s$ be an interior vertex star containing six hinges $h \in s$. Imagine cutting the vertex star (including its 6 hinges and 6 panels) out of the sheet. Let $E_{\text{star}}(s; \Pi)$ represent the elastic energy in the isolated vertex star and $E_{\text{star}}^*(s; \Pi)$ represent the minimum elastic energy after the vertex star has been relaxed using an inner solve.  

 Some hinge types may have internal strain even at their rest state (i.e. the hinge's energy function is never 0 for any angle), and this internal strain should be ignored when measuring geometric incompatibility. Thus, we define the star's reference bending energy $E_{\text{ref}}(s; \Pi)$ as the sum of its component hinges' rest energies, weighted by $\pi_{h\tau}$ at each hinge:
 $$E_{\text{ref}}(s; \Pi) \;=\; \sum_{h\in s}\;\sum_{\tau} \pi_{h\tau}\,\min_{\theta} B_\tau(\theta)$$

 The sheets used in this work have 8 interior vertices. We now define the \emph{incompatibility} of a sheet using $E_{\text{star}}^*(s; \Pi) - E_{\text{ref}}(s; \Pi)$. Since $E_{\text{ref}}$ describes the minimum possible energy, this excess term is guaranteed to be nonnegative.

 $$\mathcal{I}(\Pi) = \frac{1}{8}\sum_{s} \Big[E_{\text{star}}^*(s; \Pi) - E_{\text{ref}}(s; \Pi)\Big].$$

 The internal bending energy associated with holding a valley or mountain type fold in the flat state ($\theta = 0$) is 0.05. We call this value $E_{\text{incompatible}}$ because if a hinge expresses this much frustration, it means it has likely been assigned the wrong type. We call a vertex star $s$ \emph{incompatible} when $E_{\text{star}}^*(s; \Pi) - E_{\text{ref}}(s; \Pi) \geq E_{\text{incompatible}}$.

\subsection{Measuring Geometric Coarseness}\label{si_fourier}
Geometric coarseness can be measured using the 2D Fourier transform, which decomposes a surface into trigonometric functions of different wavelengths. An origami sheet which is coarser will have higher Fourier transform values corresponding to low-frequency, long-wavelength functions.

The Fourier transform is taken using the vertex positions $X$ of a sheet, treating the $z$-coordinate (height) as a 2D function of $x$ and $y$. The height is linearly interpolated onto a $13\times13$ grid; points outside the convex hull of vertices are assigned the height of the nearest vertex. The grid is fixed across all runs and corresponds to the bounding box of the flat sheet. A 2D Hann window is applied to reduce artifacts, and the mean value is subtracted to zero the constant term in the Fourier transform. Then the Fourier transform is applied. We split its power (squared magnitude of the Fourier transform) into six equal-width bins according to wavenumber, excluding the constant term. Finally, we define the \emph{long-wavelength power} as the fraction of the power in the two lowest bins, which correspond to wavelengths longer than about 1.6 edge lengths.

\subsection{Loss with extra penalties}\label{si_extra_penalties}

To test adding extra penalties for frustration and floppiness in the loss, we use regularizaton terms. One term penalizes the total bending energy above each hinge's reference energy, while another term penalizes the probability currently assigned to floppy ($\mathrm{F}$-type) hinges.

\begin{equation}
  \mathcal{L}_{\text{pen}}(\Pi) =
  \mathcal{L}(\Pi)
  \;+\;
  \lambda_{\text{frus}} \sum_{h}\sum_{\tau \in \mathcal{T}} \pi_{h\tau}
    \Big( B_\tau\big(\theta_h(X^*(\Pi))\big) - \min_{\theta} B_\tau(\theta) \Big)
  \;+\;
  \frac{\lambda_{F}}{N_H}\sum_{h} \pi_{hF},
  \label{eq:penalized_outer_loss}
\end{equation}
where $\lambda_{\text{frus}} = 2.97\times10^{-4}$ and $\lambda_F = 0.693$.

\end{document}